\documentclass[11pt]{article}

\usepackage[preprint]{acl}

\usepackage{times}
\usepackage{latexsym}

\usepackage[T1]{fontenc}
\usepackage[utf8]{inputenc}

\usepackage{microtype}

\usepackage{inconsolata}

\usepackage{graphicx}

\usepackage{booktabs}       
\usepackage{multirow}       
\usepackage{amsmath}        
\usepackage{amsfonts}       
\usepackage{amssymb}        
\usepackage{nicefrac}       
\usepackage{xcolor}         
\usepackage{tikz}           
\usepackage{colortbl}       
\usepackage{listings}
\usepackage[most]{tcolorbox}
\tcbuselibrary{breakable,skins,listings}
\definecolor{lightblue}{RGB}{218,232,252}

\usepackage{pifont}          
\newcommand{\cmark}{\ding{51}}
\newcommand{\xmark}{\ding{55}}

\title{ACM: Agentic Context Management for Long Horizon Tasks}

\author{
\begin{tabular}{c}
Xiaochuan Li$^{1*}$ \quad Ryan Ming$^{1*}$ \quad Meng Chu$^{1}$ \\
Shuai Shao$^{2}$ \quad Rong Jin$^{2}$ \quad Chenyan Xiong$^{1}$
\end{tabular} \\
\\
$^{1}$Carnegie Mellon University \quad
$^{2}$Meta \\
\texttt{\{xiaochu4,cx\}@andrew.cmu.edu}
}

\begin{document}
\maketitle

\begingroup
\renewcommand\thefootnote{}
\footnotetext{* Equal contribution. All experiments, data collection, and processing activities were conducted by CMU. Meta was involved solely in an advisory role and no experiments, data collection or processing activities were conducted on Meta infrastructure. Correspondence to: Xiaochuan Li <xiaochu4@andrew.cmu.edu>}
\endgroup

\begin{abstract}
Agentic tasks are inherently long-horizon and multi-turn, constantly accumulating context through interactions with the environment. Existing context compression methods inevitably incur information loss and are triggered by rigid heuristic rules, leaving them misaligned with the agent’s evolving reasoning focus. We propose \textbf{Agentic Context Management} (ACM), a framework that equips agents with purpose-built context editing tools for lossless context management. Inspired by the interaction between short-term and long-term human memory, the agent autonomously decides when to compress its context, offloads discarded content to an external memory system, and queries it on demand for later retrieval. Building on this framework, we further develop a post-training pipeline that constructs high-quality demonstrations of context management and improves model performance on both agentic search and coding tasks. Further analysis reveals that effective context management reduces peak token pressure, enables extended explorations, and yields more consistent solutions across independent trials. Code, data, and model checkpoints are available at \url{https://github.com/lixiaochuan2020/agentic-context-management}.
\end{abstract}

\section{Introduction}

\begin{table*}[t]
\centering
\normalsize
\setlength{\tabcolsep}{3pt}
\begin{tabular}{@{}lccccc@{}}
\toprule
\textbf{Method} & \textbf{Compact} & \textbf{Trainable} & \textbf{Lossless} & \textbf{Agent-init} & \textbf{Open-source Data} \\
\midrule
ACE~\citep{ace}               & \textcolor{red}{\xmark} & \textcolor{red}{\xmark} & \textcolor{red}{\xmark} & \textcolor{red}{\xmark} & \textcolor{green!60!black}{\cmark} \\
Mem1~\citep{mem1}             & \textcolor{red}{\xmark} & \textcolor{green!60!black}{\cmark} & \textcolor{red}{\xmark} & \textcolor{red}{\xmark} & \textcolor{green!60!black}{\cmark} \\
ReSum~\citep{wu2025resum}     & \textcolor{green!60!black}{\cmark} & \textcolor{green!60!black}{\cmark} & \textcolor{red}{\xmark} & \textcolor{red}{\xmark} & \textcolor{red}{\xmark} \\
ACON~\citep{acon}             & \textcolor{green!60!black}{\cmark} & \textcolor{red}{\xmark} & \textcolor{red}{\xmark} & \textcolor{red}{\xmark} & \textcolor{green!60!black}{\cmark} \\
SUPO~\citep{lu2025scaling}    & \textcolor{green!60!black}{\cmark} & \textcolor{green!60!black}{\cmark} & \textcolor{red}{\xmark} & \textcolor{red}{\xmark} & \textcolor{red}{\xmark} \\
AgentFold~\citep{ye2025agentfold} & \textcolor{green!60!black}{\cmark} & \textcolor{green!60!black}{\cmark} & \textcolor{red}{\xmark} & \textcolor{green!60!black}{\cmark} & \textcolor{red}{\xmark} \\
\midrule
\textbf{ACM (Ours)} & \textcolor{green!60!black}{\cmark} & \textcolor{green!60!black}{\cmark} & \textcolor{green!60!black}{\cmark} & \textcolor{green!60!black}{\cmark} & \textcolor{green!60!black}{\cmark} \\
\bottomrule
\end{tabular}
\caption{Comparison of context management approaches. \textbf{Compact}: actively compresses working context. \textbf{Trainable}: the management policy is learned in training. \textbf{Lossless}: raw content is preserved for later retrieval. \textbf{Agent-init}: compression is triggered by the agent itself. \textbf{Open-source Data}: training data for the context management policy is publicly released. More details can be found in Appendix~\ref{app:baseline}.}
\label{tab:related_comparison}
\end{table*}

Agentic tasks have emerged as a central challenge for LLM-powered autonomous agents~\citep{xu2024theagentcompany,xie2024osworld,swebenchpro2025,yang2024sweagent,wang2024openhands,anthropic2024claudecode,openai2025codex}. These tasks require agents to formulate adaptive plans, invoke tools~\citep{schick2023toolformer}, and adjust their actions in response to environmental feedback.
\textbf{However, the traces produced by long-horizon agentic tasks are inherently verbose and noisy}. In real-world environments, lengthy tool outputs are often interleaved with failed attempts and redundant observations. When combined with the agent’s own reasoning traces, they accumulate into histories that exceed an agent’s effective context capacity, even when the underlying model supports nominal context windows of millions of tokens~\citep{gemini15,anthropic2025sonnet4,openai2025gpt41}.

Prior work has explored several directions to mitigate this limitation. Long-context pretraining extends the window but exhibits measurable degradation~\citep{liu2024lostmiddle,hsieh2024ruler,bai2024longbench,hong2025context}. Hybrid attention reduces the cost of processing long inputs but still remains fundamentally bounded by the context window~\citep{dao2024transformers, deepseekv32, lenz2025jamba}. Context-compression pipelines — which truncate, summarize, or re-render histories into denser formats — represent a promising direction~\citep{wei2025deepseek, acon}. However, these approaches control compression timing through forced, hand-crafted external monitors, relying on heuristic rules that are not well aligned with the model's own reasoning process.

In this paper, we propose a framework that enables model-intrinsic and lossless context management. Specifically, by equipping the agent with a set of well-designed memory tools, we allow the agent itself to decide \textbf{when} and \textbf{how} to manage its context: identifying irrelevant information, summarizing and offloading it to external memory, and querying it on demand. Our design draws inspiration from the separation between short-term and long-term memory in human cognition~\citep{atkinson1968memory,packer2023memgpt} — in-context messages serve as a compact working memory buffer focused on reasoning, while an external store acts as long-term memory ready for future retrieval. This mechanism enables the agent to expand or contract its effective context as its understanding of task progress evolves.

Building on this framework, we further develop an efficient post-training pipeline to help the model internalize context management ability.  We adopt a teacher--student on policy framework~\citep{hinton2015distilling, lu2025onpolicydistillation} with dual constraints. In one direction, the student performs rollouts \emph{without} context management, and the teacher reviews the resulting trajectories to identify where context management should be inserted, e.g., when the model is stuck in a dead-end loop. In the other direction, the student generates another set of rollouts \emph{with} full access to context management tools, and the teacher identifies where the context management should \emph{not} have been called---replacing it with either a commitment to an answer or a deeper search action. The dual constraints teach the student agent the accurate timing of context management. We then prompt the student to resume and complete the task from the point where the teacher provides feedback, while using the teacher’s assessments of the student’s trajectories as soft supervision signals for training. Using this dual-constraint pipeline, we improve Qwen3.5-9B’s search and coding performance over the ReAct baseline by 27\% on BrowseComp-Plus~\citep{chen2025BrowseCompPlus}, 16\% on DeepSearchQA~\citep{gupta2026deepsearchqa}, and 8\% on SWE-Bench Verified~\citep{jimenez2024swe}. Analysis reveals that ACM reduces peak token usage by around 20\%, increases tool call frequency, and extends the test-time exploration turns. These gains translate into more consistent solutions across independent trials. In summary, our contributions are as follows:

\begin{itemize}
\item We introduce \emph{agentic context management}, a paradigm in which the agent autonomously decides \textbf{when} and \textbf{how} to manage its own context.
\item We propose an efficient post-training pipeline that internalizes context management ability into the model itself.
\item We demonstrate that effective context management reduces peak token pressure, extends test-time exploration turns, and improves solution consistency across independent trials.

\end{itemize}

\section{Related Works}

\begin{figure*}[h]
    \centering
    \includegraphics[width=0.90\linewidth]{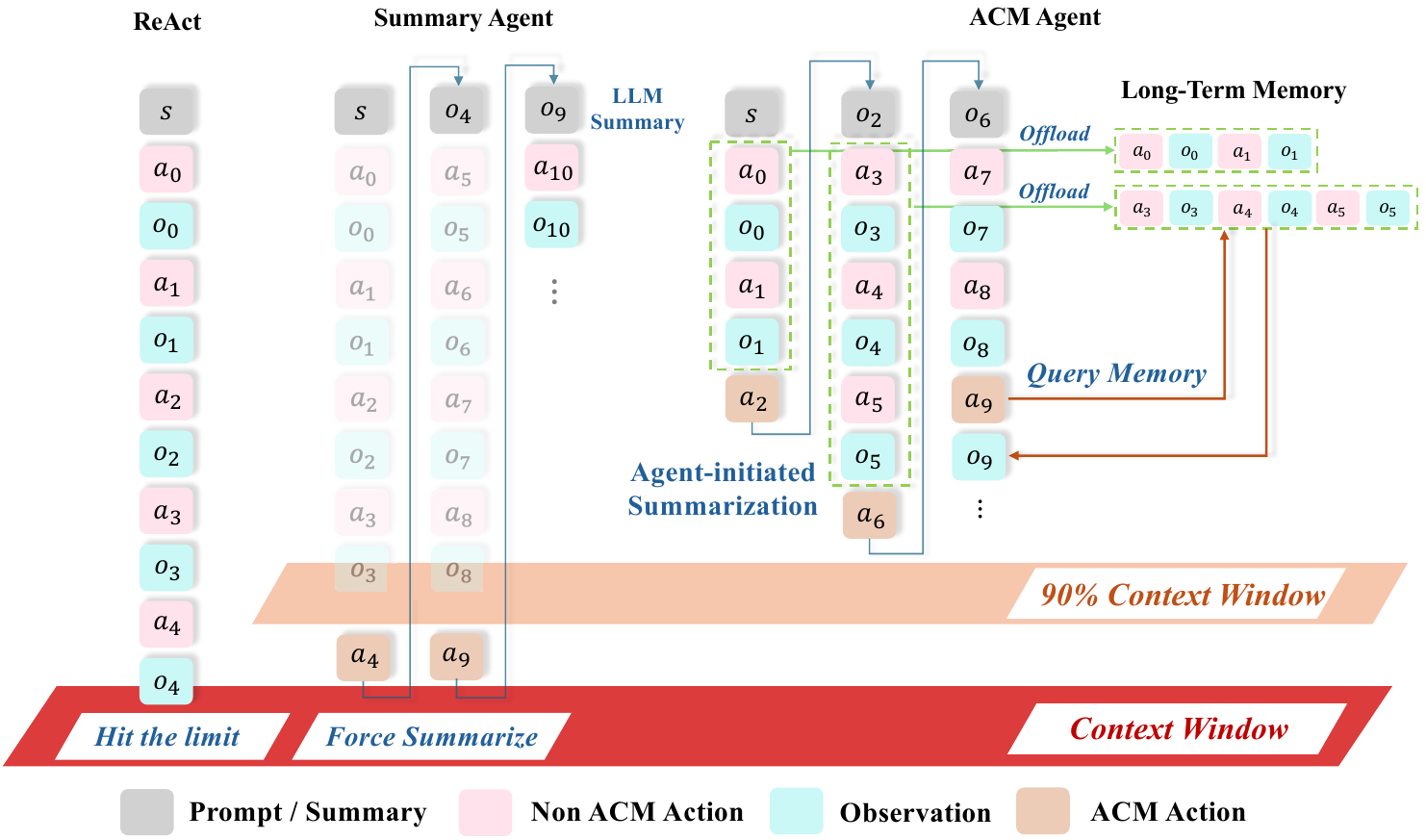}
    \caption{Overview of our ACM Framework. ReAct eventually hits the context limit, while the Summary Agent is forced to compress whenever usage exceeds a predefined threshold (e.g., 90\% of the context in the figure) and discards the original messages. The ACM agent autonomously decides \emph{when} to manage its context losslessly.}
    \label{fig:acm_pipeline}
    \vspace{-0.5em}
\end{figure*}

\subsection{Heuristic Context Compression}

Heuristic Context Compression reduces context length through external compression modules or hand-designed discarding rules that operate outside the agent's decision process. ReSum~\citep{wu2025resum} is among the first frameworks to adopt fixed-timing compression for agentic search tasks. ACON~\citep{acon} trains a compressor that replaces prior histories with a condensed summary, an approach also adopted in Claude's Automatic Context Compression~\citep{anthropic2024claudecode}. COMPASS~\citep{compass} delegates compression to a Meta-Thinker agent that supplies compact contexts to the main agent. DeepSeek-V3.2~\citep{deepseekv32} studies manually designed strategies such as full-history summarization and fixed-ratio truncation, and shows that accuracy continues to improve with step count. MemAgent~\citep{memagent} processes inputs chunk by chunk, discarding earlier content while retaining only last-round memory and the question. SideQuest~\citep{sidequest} takes an orthogonal infrastructure-level approach by managing the KV cache directly during long-horizon agentic reasoning. \citet{lu2025scaling} (SUPO) and \citet{sun2025scaling} both use reinforcement learning~\citep{shao2024deepseekmath} to teach the agent to summarize or fold context. Collectively, these approaches demonstrate the utility of context reduction, but they rely on external modules or fixed heuristics rather than on decisions made by the agent during reasoning. Our work instead treats context management as an explicit agent action during task execution. AgentFold~\citep{ye2025agentfold} is closely related to our work, but its data-generation pipeline is not publicly available. We complement this by open-sourcing a complete data generation pipeline that enables efficient post-training without heavy reinforcement learning.

\subsection{Memory-Augmented Context Evolution}

Memory-Augmented approaches~\citep{singh2025agentic,packer2023memgpt,park2023generative,wang2023voyager,xu2025mem,fang2025memp} maintain an external memory that is updated continuously---typically through reflection, distillation, or optimization---so that the working context remains concise while accumulated knowledge is stored elsewhere. MIPRO~\citep{mipro} jointly optimizes instructions and demonstrations across multi-stage LM programs. Dynamic Cheatsheet~\citep{dynamiccheatsheet} learns a reusable note at test time that records useful strategies. GEPA~\citep{gepa} evolves prompts and contexts through reflective Pareto search. Agentic Context Engineering~\citep{ace} treats the context itself as an evolving artifact refined through self-improvement loops. Mem1~\citep{mem1} consolidates trajectories into a compact internal memory state that is updated each turn.  These approaches accumulate knowledge \emph{across tasks} but do not compress the working context within a single episode. ACM, by contrast, is a purely \emph{per-question} method that dynamically compresses and retrieves context through explicit tool calls during reasoning. Table~\ref{tab:related_comparison} summarizes the key differences between our method and representative prior work.

\section{Agentic Context Management Framework}\label{section:acm_framework}

\begin{figure*}[t]
    \centering
    \includegraphics[width=0.99\linewidth]{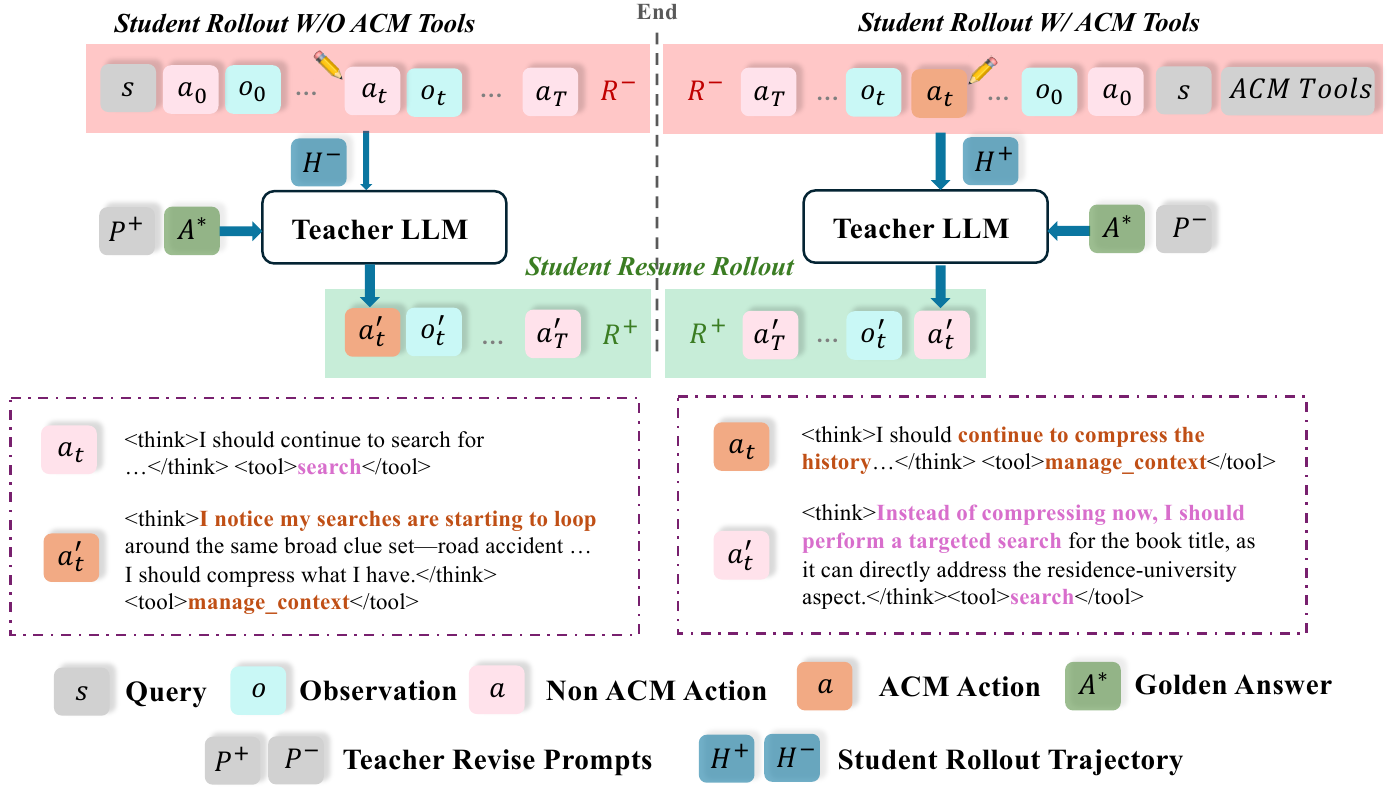}
    \caption{Overview of the dual-constraint training data generation pipeline. A student agent completes task rollouts both with and without context management tools. A teacher model reviews each trajectory against the reference answer and either injects ACM action or replaces non-ACM actions. }
    \label{fig:data_pipeline}
\end{figure*}

\paragraph{Formulation}

Let $s$ denote the system prompt, $a_t$ an agent action (reasoning content plus tool calls), and $o_t$ the corresponding environment response at turn $t$. The agent $\pi_{\theta}$ conditions on the accumulated history $H_t = \bigl\{s,\,(a_1, o_1),\,\dots,\,(a_{t-1}, o_{t-1})\bigr\}$ to produce:
\begin{equation*}
  a_t \sim \pi_{\theta}(\cdot \mid H_t),
  \qquad
  o_t \sim \pi_{\gamma}(\cdot \mid H_t;\, a_t),
\end{equation*}
where $\pi_{\gamma}$ is the model that returns environment response via tool results. The interaction terminates when the agent selects the \texttt{finish} action $a_T$ or its context window reaches the limit.

\paragraph{Summary Agent.} In the summary-agent paradigm, an external monitor triggers the compression action $a_{\text{sum}}$ when context usage exceeds a predefined threshold. The environment returns a summary $o_{\text{sum}} \sim \pi_{\gamma}(\cdot \mid H_t;\, a_{\text{sum}})$, then the agent discards all prior messages and continues reasoning over the updated history $H' = \{s,\, o_{\text{sum}}\}$.

\paragraph{ACM Agent.} We draw inspiration from the interaction between short-term and long-term memory in human cognition: people keep immediately relevant information in working memory while offloading less immediate details into external persistent records, and retrieve them later when the task requires. We introduce only \textbf{two} context management tools to enable the agent to mimic the human memory mechanism: \texttt{manage\_context}, which compresses previous turns into a concise summary and offloads the raw messages to an external file on disk; and \texttt{query\_memory}, which allows the agent to query the stored raw messages to retrieve information precisely.

The overall mechanism of ACM, along with a comparison to the ReAct and the Summary agent, is illustrated in Figure~\ref{fig:acm_pipeline}. When the agent decides to manage its context, it invokes \texttt{manage\_context} (action $a_2, a_6$ in Figure~\ref{fig:acm_pipeline}) to compress all messages up to the previous summary boundary using a summarizer LLM. Crucially, the original messages are not discarded but saved to the agent's external workspace. Each summary is assigned a unique identifier that maps the summary to the corresponding raw messages in external memory. When the agent needs to revisit earlier content, it invokes \texttt{query\_memory} (action $a_9$ in Figure~\ref{fig:acm_pipeline}) with a specified identifier. A querier LLM receives the query along with the raw messages mapped by that identifier, then returns the information related to the query as a tool result.

ACM has two key properties that distinguish it from prior summary-based agents. \textbf{1) Information compression is lossless}: all discarded messages are preserved in external storage and are available for the agent to revisit at any time, while the working context stays short and clean. \textbf{2) Context management is agent-initiated}: the agent can invoke compression at any point during reasoning process, rather than relying on a fixed schedule or an external trigger. This enables context management to follow the agent's evolving reasoning state and task progress. By allowing compression at any point before the history reaches peak length, the design also alleviates peak token usage pressure.

\begin{table*}[t]
\centering
\small
\begin{tabular}{@{}lccccccccc@{}}
\toprule
& \multicolumn{3}{c}{\textbf{BrowseComp-Plus}} & \multicolumn{3}{c}{\textbf{DeepSearchQA}} & \multicolumn{3}{c}{\textbf{SWE-Bench Verified}} \\
\cmidrule(lr){2-4} \cmidrule(lr){5-7} \cmidrule(lr){8-10}
\textbf{Method} & Pass@1 & Tools & Peak Tok. & Pass@1 & Tools & Peak Tok. & Pass@1 & Tools & Peak Tok. \\
\midrule
\rowcolor{lightblue} \multicolumn{10}{l}{\textit{Frontier Models}} \\
\quad Qwen3.5-397B-A17B & 0.653 & 15.6 & 51K & 0.710 & 28.3 & 47K & 0.682 & 58.9 & 38K \\
\quad Gemini3-Flash             & 0.733 & 22.9 & 72K & 0.619& 54.3 & 121K & 0.732 & 66.7 & 80K \\
\midrule
\rowcolor{lightblue} \multicolumn{10}{l}{\textit{ReAct}} \\
\quad Qwen3.5-9B            & 0.570 & 19.5 & 63k & 0.367 & 47.4 & 46k & 0.489 & 74.7 & 59k \\

\addlinespace
\rowcolor{lightblue} \multicolumn{10}{l}{\textit{Summary Agent + Qwen3.5-9B}} \\
\quad ReSum            & 0.608 & 24.7 & 68k & 0.371 & 48.6 & 79K & 0.475 & 75.2 & 61K \\
\quad ACON            & 0.614 & 28.2 & 65k & 0.380 & 51.3 & 54K & 0.480 & 76.1 & 57K \\
\rowcolor{lightblue} \multicolumn{10}{l}{\textit{Memory Agent + Qwen3.5-9B}} \\
\quad ACE            & 0.589 & 19.8 & 71k & 0.352 & 48.2 & 70K & 0.494 & 75.6 & 65K \\
\midrule
\rowcolor{lightblue} \multicolumn{10}{l}{\textit{ACM Agent + Qwen3.5-9B}} \\
\quad Base            & 0.635 & 30.8 & 59k & 0.405 & \textbf{88.7} & 42K & 0.508 & 77.6 &  \textbf{46K} \\
\quad ACM-Post-Trained     & \textbf{0.727} & \textbf{46.2} & \textbf{54k} & \textbf{0.425} & 58.8 & \textbf{41K} & \textbf{0.530} & \textbf{79.3} & 50K \\
\bottomrule
\end{tabular}
\caption{Main results on BrowseComp-Plus, DeepSearchQA, and SWE-Bench Verified. Pass@1 reports accuracy. Tools is the average number of tool calls per episode. Peak Tok.\ is the average peak token count across episodes.}
\label{tab:main_results}
\end{table*}

\section{Training Data Generation}
\label{sec:data_generation}

As discussed in Section~\ref{tool_use_decomp} and supported by \citet{ye2025agentfold,lu2025scaling}, even frontier models struggle to determine the appropriate timing for context management. To address this gap, we design a teacher-guided data generation pipeline with dual constraints that is both easy to scale and capable of producing high-quality management demonstrations. We reuse the formulation in Section~\ref{section:acm_framework}.

\subsection{Teacher-Guided Annotation}

Our pipeline employs a teacher--student framework with dual constraints and operates in two phases, as illustrated in Figure~\ref{fig:data_pipeline}.

\paragraph{Phase 1: Student Rollout.}
A student model completes the task under two conditions---\emph{with} and \emph{without} access to context management tools---producing trajectories denoted $H^{+}$ and $H^{-}$, respectively. $H^{+}$ captures the student's untrained usage behavior of the context management tools, while $H^{-}$ reflects its ordinary exploration behavior. In Figure~\ref{fig:data_pipeline}, the $H^{-}$ rollout is shown on the left, where the student starts from the system prompt $s$ alone without ACM tools, whereas the $H^{+}$ rollout is shown on the right, starting from $s$ together with the ACM tools.

\paragraph{Phase 2: Teacher Annotation.}
A \emph{teacher} model receives one of two guided instruction prompts, $P^{+}$ or $P^{-}$: 
1) teacher using $P^{+}$ demonstrates \emph{when to use} the context management tools while 2) teacher using $P^{-}$ demonstrates \emph{when not to} call them. The teacher is additionally provided with the corresponding student trajectory ($H^{+}$ or $H^{-}$) and the reference answer $A^{*}$. It then reviews the trajectory and produces annotations under two complementary constraints:

\begin{itemize}
    \item \textbf{Injection on $H^{-}$} (where to \emph{add} context management). Given $P^{+}$, the teacher identifies turns at which context management would be beneficial---specifically, points where the student begins querying redundant topics, enters unproductive loops, or has accumulated sufficient context to warrant compression. At each such turn $t$, the teacher uses a context management tool call $a_t'$ accompanied by a reasoning trace that justifies the compression.
    \item \textbf{Refinement on $H^{+}$} (where to \emph{remove} context management). Given $P^{-}$, the teacher identifies turns at which the student's context management calls are premature or unnecessary. At each such turn $t$, the student has typically either gathered sufficient information but failed to synthesize a final answer, or overlooked key evidence in the retrieved documents that warrants deeper exploration. The teacher replaces the inappropriate context management call $a_t$ with a more productive action $a_t'$---such as searching for additional evidence, opening a relevant document, or committing to an answer---accompanied by a reasoning trace.
\end{itemize}

In both cases, the student's original action $a_t$ is replaced with the teacher-annotated action $a_t'$, and the student rollout resumes from $a_t'$. 

We then train the student using on-policy distillation~\citep{lu2025onpolicydistillation}. A stronger teacher from the same model family annotates each student-generated assistant token with a soft next-token distribution. In practice, we retain the teacher probabilities for the top-$K$ tokens, with $K=20$. The student is optimized to match these teacher distributions over all assistant-token positions in the rollout:
\begin{equation*}
\begin{aligned}
\mathcal{L}_{\mathrm{ACM}}(\theta)
={}& -\,\mathbb{E}_{\tau \sim \pi_{\theta}}
\Bigg[
\sum_{t \in \mathcal{T}_{a}(\tau)}
\\[-2pt]
&\sum_{v \in \mathcal{V}}
p_{\mathrm{T}}\!\left(v \mid s; h_{<t}\right)
\log \pi_{\theta}\!\left(v \mid s; h_{<t}\right)
\Bigg].
\end{aligned}
\label{eq:acm_loss}
\end{equation*}
where $\tau$ is a trajectory sampled from the student policy, $\mathcal{T}{a}(\tau)$ denotes the set of assistant-token positions, and $\mathcal{V}$ contains the teacher’s top-$K$ candidate tokens at position $t$. The distribution $p_{\mathrm{T}}(\cdot \mid s;h{<t})$ denotes the teacher probabilities restricted and renormalized over $\mathcal{V}$, while $\pi_{\theta}(\cdot \mid s; h_{<t})$ denotes the student’s next-token distribution. Here, $h_{<t}=(a_{1},o_{1},\dots,a_{t-1},o_{t-1})$ represents the interleaved history of preceding actions and tool observations. The loss is applied to all student-generated assistant tokens, while system-prompt, user-input, and tool-output tokens are masked out. Under this objective, the student jointly learns when to invoke context management and when to refrain from doing so because a search, retrieval, or commit-to-answer action is more appropriate.

\subsection{Quality Filtering}

We apply two filtering mechanisms to ensure data quality: \textbf{1) Rejection sampling}~\citep{yuan2023rft,touvron2023llama2}\textbf{:} we retain only trajectories in which the student fails to complete all trials successfully. This ensures that the student learns from the teacher’s behavior on genuinely challenging problems. \textbf{2) Content filters:} filters are applied to verify that the teacher's reasoning traces do not leak information from the reference answer $\mathcal{A}^{*}$. The teacher's annotations must explain \emph{why} compression is warranted or unnecessary at turn $t$---citing cues such as redundant queries, cyclic exploration patterns, or sufficient evidence to commit to an answer---without revealing the target answer itself. The constrained training data encourages the model to recognize compression-worthy patterns from the trajectory structure rather than memorizing answer-dependent cues. Finally, to stabilize training, we resample trajectories from the student's original rollouts, in a manner similar to self-distillation~\citep{zelikman2022star}, and mix them with the teacher-annotated data.

\begin{table*}[t]
    \centering
    \small
    \begin{tabular}{@{}lccccccccc@{}}
    \toprule
    & \multicolumn{3}{c}{\textbf{BrowseComp-Plus}} & \multicolumn{3}{c}{\textbf{DeepSearchQA}} & \multicolumn{3}{c}{\textbf{SWE-Bench Verified}} \\
    \cmidrule(lr){2-4} \cmidrule(lr){5-7} \cmidrule(lr){8-10}
    \textbf{Method} & Pass@1 & Tools & Peak Tok. & Pass@1 & Tools & Peak Tok. & Pass@1 & Tools & Peak Tok. \\
    \midrule
    Qwen3.5-9B              & 0.635 & 30.8 & 59k & 0.405 & 88.7 & 42K & 0.508 & 77.6 & 46K \\
    \quad + GPT5.5 Distill   & 0.623 & 26.4 & 62k & 0.381 & 49.6 & 53K & 0.542 & 58.3 & \textbf{45K} \\
    \quad + ACM              & 0.727 & 46.2 & \textbf{54k} & \textbf{0.425} & 58.8 & \textbf{41K} & 0.530 & 79.3 & 50K \\
    \quad + Both             & \textbf{0.734} & \textbf{37.6} & 59k & 0.413 & \textbf{62.5} & 50K & \textbf{0.564} & \textbf{88.1} & 57K \\
    \bottomrule
    \end{tabular}
    \caption{Ablation of distillation and ACM training on Qwen3.5-9B. Pass@1 reports accuracy. Tools is the average number of tool calls per episode. Peak Tok.\ is the average peak token count across episodes.}
    \label{tab:ablation_distill}
\end{table*}

\section{Experiments}

\subsection{Experimental Setup}

\paragraph{Tasks and Datasets.}
We evaluate our method on three long-horizon agentic benchmarks: BrowseComp-Plus~\citep{chen2025BrowseCompPlus}, DeepSearchQA~\citep{gupta2026deepsearchqa}, and SWE-Bench Verified~\citep{jimenez2024swe}. Simple tasks rarely require context management, as they are typically solved before substantial context pressure arises. For BrowseComp-Plus, we use 680 examples for training and 150 for evaluation. DeepSearchQA is used exclusively as an out-of-domain evaluation benchmark with access to a live web search engine. For SWE-Bench Verified, we use SWE-Gym~\citep{pan2024training} as the training dataset.

\paragraph{Data Generation.}
We use Qwen3.5-9B~\citep{yang2025qwen3} as the student rollout model, as well as the summarizer and querier, because it can generate sufficiently long and coherent trajectories to provide meaningful demonstrations of context management. Substantially smaller models often lose coherence after only a few turns, producing trajectories of limited value for learning effective compression behavior. \textbf{We additionally compare against Qwen3-4B-Thinking in Appendix~\ref{app:case-study-qwen3-4b} to demonstrate the effect of model scale.} We use Qwen3.5-397B-A17B as the teacher model and perform on-policy distillation for three epochs. We also open-source the student’s four rollout trials and the corresponding teacher annotations from each epoch.

\paragraph{Baselines.}
We compare ACM against three agent frameworks: (1)~\textbf{ReAct}~\citep{yao2022react}, the standard reasoning-and-acting agent without any context management; (2)~\textbf{Summary Agent}~\citep{wu2025resum, acon}, which triggers summarization when context usage exceeds a fixed threshold; and (3) ~\textbf{Memory Agent}~\citep{ace}, which accumulates experiences from previous rollouts but does not dynamically manage its intra-trajectory context. We also compare our method against two stronger models.

\begin{figure}[t]
    \centering
    \includegraphics[width=\columnwidth]{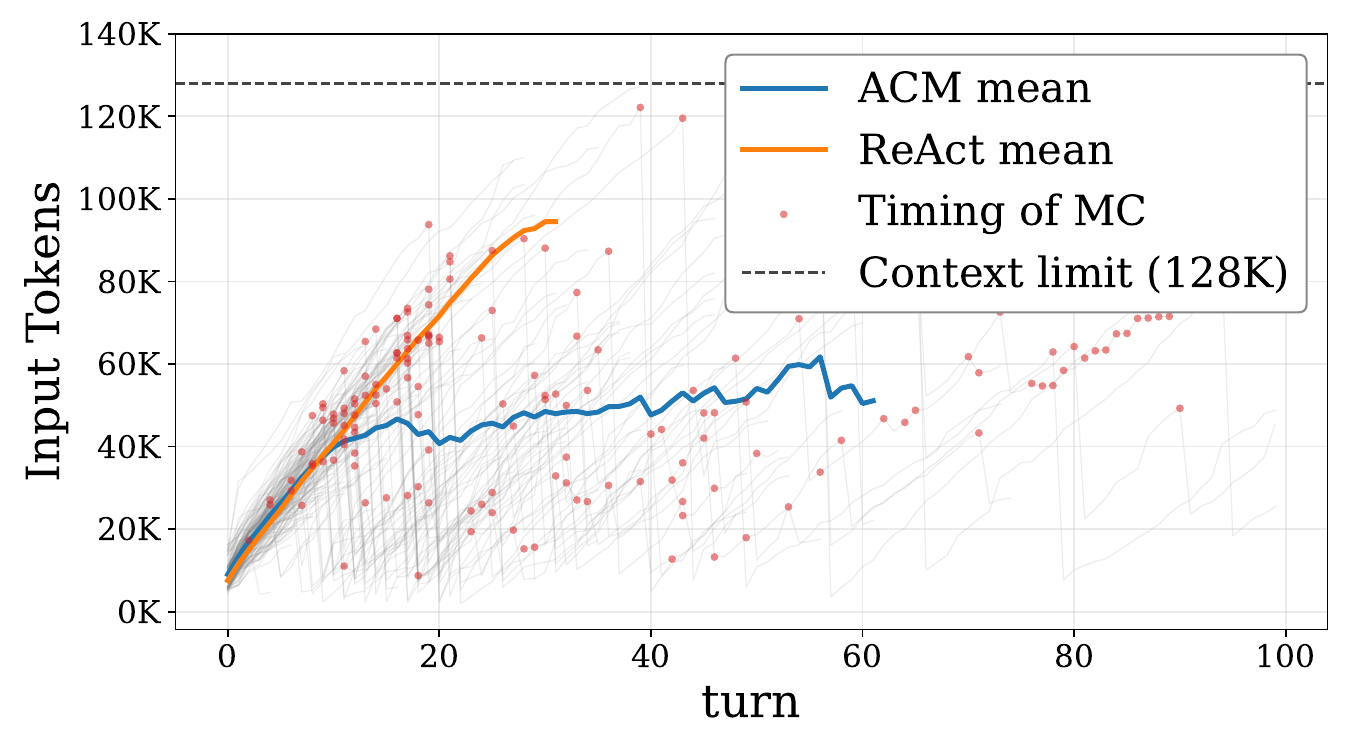}
    \caption{Input token count over interaction turns for ACM and ReAct agents. Gray curves show individual ACM trajectories; red dots mark context management calls. Yellow and blue curves denote the population average for ReAct and ACM, respectively.}
    \label{fig:tokens_vs_turns}
\end{figure}

\subsection{Main Results}

Table~\ref{tab:main_results} presents the main results. We find that by simply equipping the agent with our ACM framework, the performance of agent already surpasses all baselines, demonstrating the effectiveness of agent-initiated context management. Post-training on our high-quality context management data further improves performance, yielding a 27\% relative gain on BrowseComp-Plus and nearly matching open-source models that are 40$\times$ larger.

We also observe a positive correlation between Pass@1 and the number of tool calls. Unlike strong frontier models, which achieve high accuracy with a small number of tool calls, smaller agent models rely more on exploration to solve the problem, and context management enables them to explore effectively. Furthermore, peak token usage decreases dramatically under the ACM framework, especially compared with the Summary Agent. Therefore, the ACM framework reduces both the model's reasoning burden and the server's KV-cache overhead.

\subsection{Behavior study}

\paragraph{Context Growth Dynamics.} Figure~\ref{fig:tokens_vs_turns} compares the context growth of ReAct and ACM agents on BrowseComp-Plus. We can observe 2 key findings:  \textbf{1) ACM agents learn to compress context proactively:} the characteristic sawtooth pattern shows that compression is triggered well before the context limit, driven by the agent's own reasoning state. \textbf{2) The payoff of context management is substantial:} By keeping the context compact, ACM substantially slows context growth while enabling more exploratory turns. As a result, the agent can continue reasoning and interacting for significantly longer before reaching the context limit. This advantage is particularly important for challenging questions that require extended, multi-step exploration and reasoning.

\begin{figure}[t]
    \centering
    \includegraphics[width=\columnwidth]{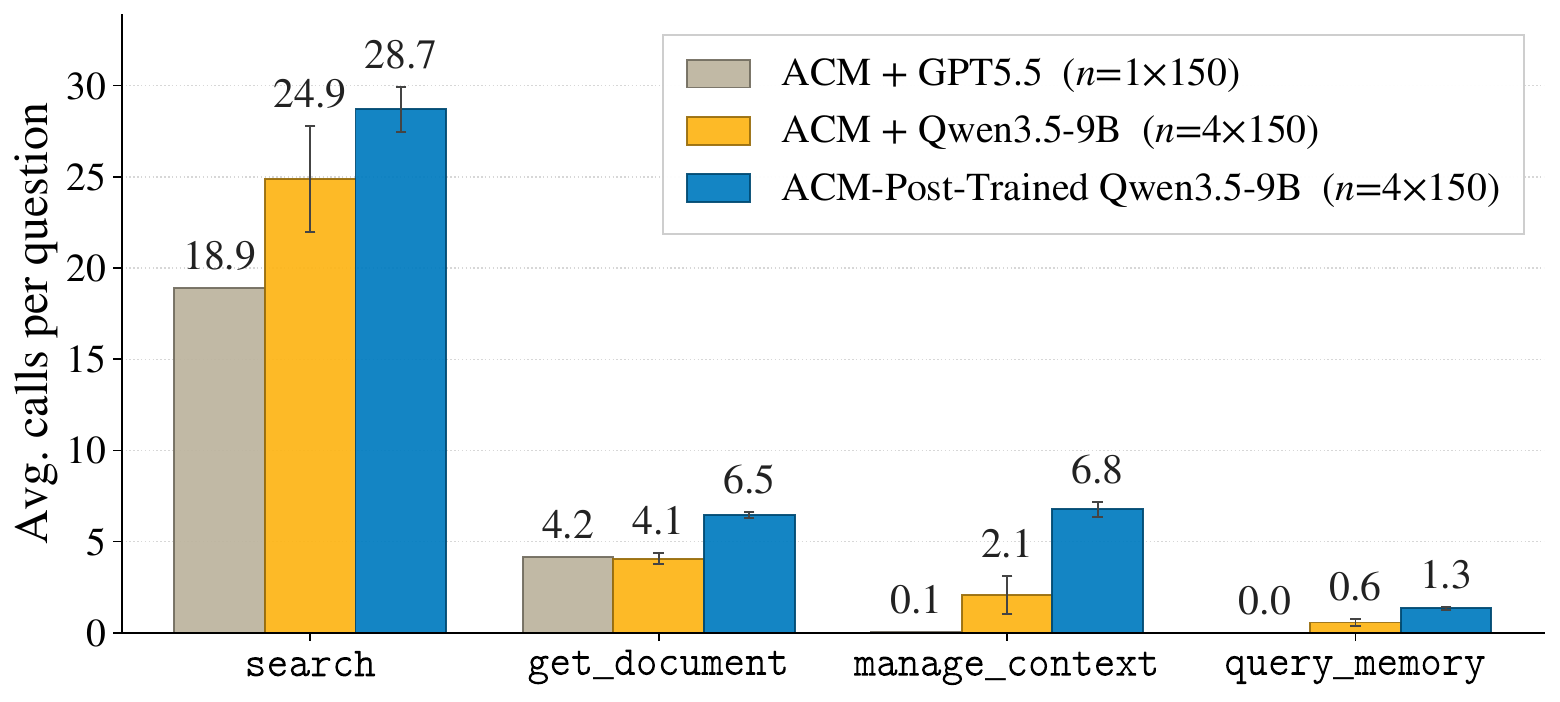}
    \caption{Per-tool call frequency on BrowseComp-Plus for GPT-5.5, Qwen3.5-9B, and our post-trained Qwen3.5-9B under the ACM framework.}
    \label{fig:tool_call_freq}
\end{figure}

\begin{figure*}[t]
    \centering
    \includegraphics[width=1.5\columnwidth]{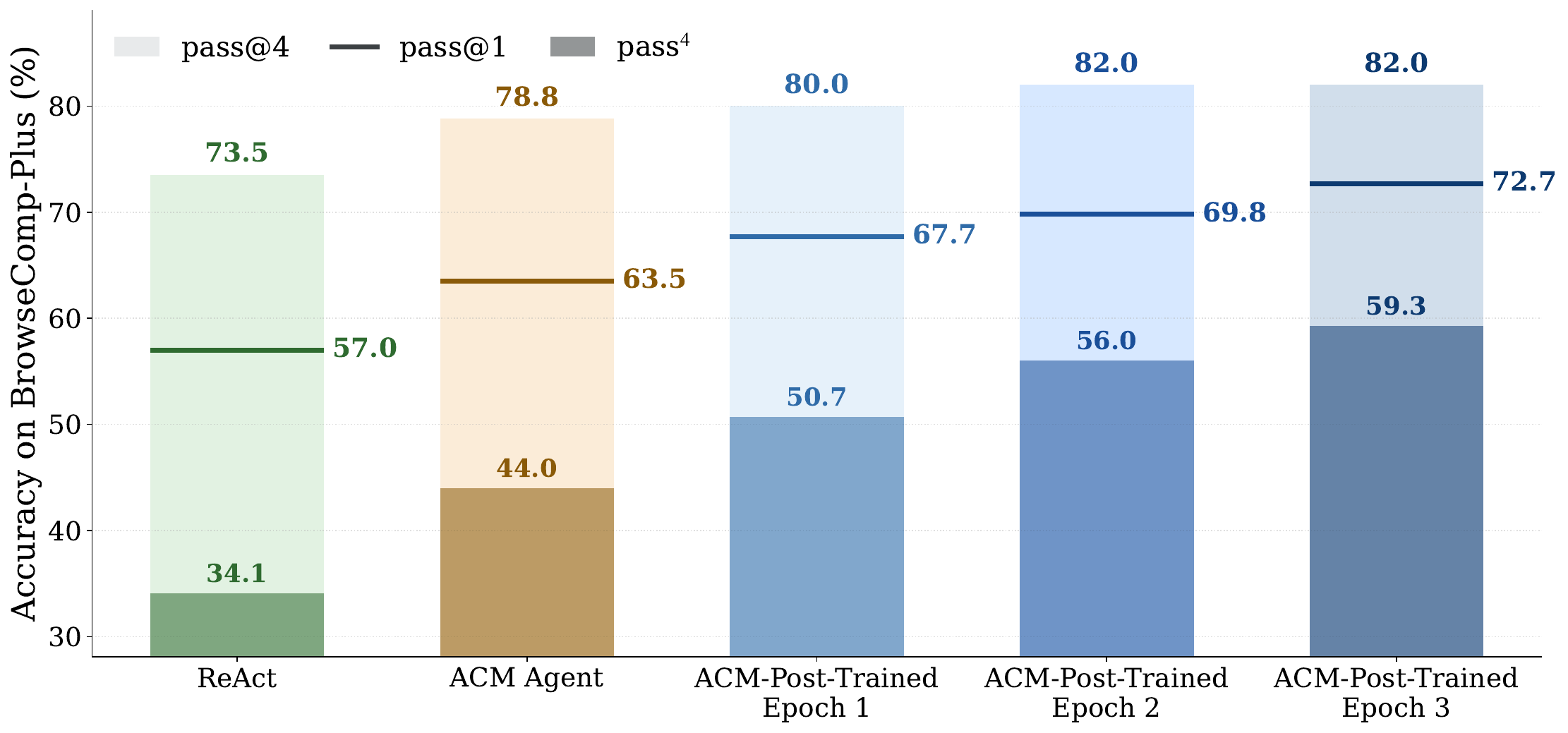}
    \caption{Accuracy of Qwen3.5-9B on BrowseComp-Plus under ReAct, ACM, and three epochs of ACM post-training. Light bars denote pass@4, dark bars denote pass$^4$, and the solid line indicates average pass@1 performance. ACM consistently improves all three metrics.}
    \label{fig:acm_vs_react}
\end{figure*}

\paragraph{Tool Usage Decomposition.}\label{tool_use_decomp}
Figure~\ref{fig:tool_call_freq} breaks down the per-tool call frequency across agents. Under the ACM framework, GPT-5.5 rarely invokes context management, making near-zero calls to both \texttt{manage\_context} and \texttt{query\_memory}. This observation suggests that even strong models may lack the proactivity to manage their own context without dedicated training. It also motivates curating high-quality trajectories from the student itself rather than relying solely on teacher distillation, since many tasks are easy enough for the teacher to solve without using context management, resulting in too few relevant behaviors for effective supervision. ACM-Post-Trained achieves the highest frequency of context management calls. Crucially, the active compression also unlocks more exploration: ACM-Post-Trained issues the most \texttt{search} and \texttt{get\_document} tool calls among all agents, enabling it to explore a broader set of reasoning paths.

\definecolor{mc}{HTML}{D9534F}    
\definecolor{qm}{HTML}{2D8C3C}    
\definecolor{gd}{HTML}{B68A00}    
\definecolor{tag}{HTML}{7A6230}   

\newcommand{\step}[1]{%
  \tikz[baseline=(c.base)]\node[circle, draw=tag, fill=tag!12,
        line width=0.45pt, inner sep=1pt, minimum size=11pt,
        font=\scriptsize\bfseries](c){#1};%
}

\begin{figure*}[!t]
\centering
\fbox{\begin{minipage}{0.97\linewidth}\small
\textbf{Question (BCP qid 347, 5-constraint multi-hop).} ``Restaurant mentioned in the acknowledgments of a UC dissertation (2010--2013); author has a B.Tech (IIT BHU) and a UCLA master's, co-authored papers in 2020 and 2020--2022; restaurant founded 1980--1988.''
\quad\textbf{Gold answer:} \textsc{California Pizza Kitchen} (A.~Jain, \emph{New Frontiers in Secure Computation}, UCLA 2012).
\end{minipage}}

\vspace{2pt}

\begin{minipage}[t]{0.40\linewidth}
\vspace{0pt}%
\includegraphics[width=\linewidth]{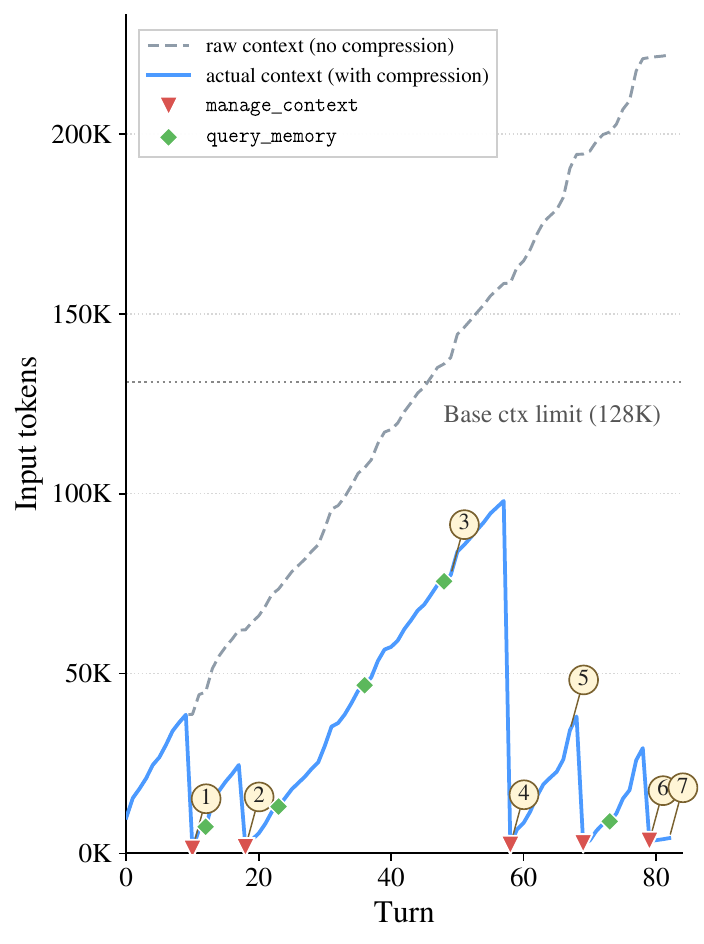}
\end{minipage}\hfill
\begin{minipage}[t]{0.58\linewidth}
\vspace{0pt}%
\renewcommand{\arraystretch}{1.05}
\scriptsize
\begin{tabular}{@{}p{0.6em}p{0.93\linewidth}@{}}
\toprule
\multicolumn{2}{@{}l}{\textbf{Model output at key turns (verbatim, abbreviated)}}\\
\midrule

\step{1} & \textbf{turn 10.} \emph{``my context token count is at 54{,}932.
  I need to find the specific dissertation efficiently.''}\par
  \verb|→ |\textcolor{mc}{\texttt{manage\_context}}
  \textcolor{mc}{\scriptsize[21 msgs $\to$ summary\_1]}\\

\step{2} & \textbf{turn 18.} \emph{``context is at 40{,}982 tokens,
  approaching half of 81{,}920 -- I need to manage context soon.''}\par
  \verb|→ |\textcolor{mc}{\texttt{manage\_context}}
  \textcolor{mc}{\scriptsize[15 msgs $\to$ summary\_2]}\\

\step{3} & \textbf{turn 49.} \emph{``Current context at 92K
  ($\sim$70\% usage). The \textcolor{qm}{\textbf{query\_memory}}
  result provides a significant lead -- mentioning `Rukmava Chatterjee'
  as a UC student\dots However, I need to verify this.''}\\

\step{4} & \textbf{turn 58.} \emph{``Now at 114{,}364 of 131{,}072
  ($\sim$87\% usage), only 16{,}606 tokens remaining.''}\par
  \verb|→ |\textcolor{mc}{\texttt{manage\_context}}
  \textcolor{mc}{\scriptsize[\textbf{78 msgs} $\to$ summary\_3,
  raw=158K vs actual=2.3K]}\\

\step{5} & \textbf{turn 67.} \emph{``Found docid 46759 --
  \textbf{`New Frontiers in Secure Computation' by Abhishek Jain},
  UCLA 2012. This could match the profile.''}\par
  \verb|→ |\textcolor{gd}{\texttt{get\_document(docid="46759")}}\\

\step{6} & \textbf{turn 79.} \emph{``Found docid 85750:
  \textbf{`Statistical Zaps and New OT Protocols' -- EUROCRYPT 2020},
  co-authored by Jain. This confirms criterion (3).''}\par
  \verb|→ |\textcolor{mc}{\texttt{manage\_context}}
  \textcolor{mc}{\scriptsize[20 msgs $\to$ summary\_4]}\\

\step{7} & \textbf{turn 82.} \emph{``Context is at 20{,}325 tokens,
  very low compared to the 131{,}072 limit. I have enough
  information to provide the final answer.''}\par
  \textbf{Answer:} \textsc{California Pizza Kitchen} (95\% conf.)\\

\bottomrule
\end{tabular}

\vspace{3pt}
\small
\textbf{Total tool use:} 63\,$\times$\,\texttt{search},
9\,$\times$\,\texttt{get\_document},
\textcolor{mc}{7\,$\times$\,\texttt{manage\_context}},
\textcolor{qm}{5\,$\times$\,\texttt{query\_memory}}; 83 turns.
\textbf{Peak} actual / raw context = 98K / 222K
($\Rightarrow$\,compression saved $\sim$124K vs the 128K base limit).
\textbf{Base model:} 0/4 success on this question.
\end{minipage}

\caption{%
\textbf{Trajectory of our ACM-Post-Trained Model on a 5-constraint
multi-hop question.} The model self-monitors its context, alternates
\textcolor{mc}{\texttt{manage\_context}} ($\blacktriangledown$) with
\textcolor{qm}{\texttt{query\_memory}} ($\blacklozenge$) probes, and
terminates only after confirming the working context is small enough.
Without compression the trajectory would cross the 128K base-model
limit at turn 47 (dashed line); the base model in fact gives up on
this question in all 4 runs.}
\label{fig:case-study-qid347}
\end{figure*}

\section{Ablation Study}

\subsection{Study of Pass@K}

We examine how agentic context management affects Pass@K~\citep{brown2024monkeys,snell2024scaling} and Pass$^K$~\citep{yao2024tau} with $K{=}4$, treating Pass@4 as a proxy for the model's \emph{capability boundary} and Pass$^4$ as a measure of \emph{consistency}. As shown in Figure~\ref{fig:acm_vs_react}, under ReAct the two diverge sharply: Pass$^4$ is much lower, reflecting the instability of long-horizon reasoning as accumulated context noise degrades decision quality.

Post-training on context-management data narrows this gap. While Pass@4 improves modestly, Pass@1 and Pass$^4$ increase substantially, suggesting that the main benefit of context management lies not only in expanding the set of problems the model can solve, but also in making correct solutions more reliable. A clean and well-organized context enables the agent to produce correct answers consistently across independent trials. Performance also improves steadily over three epochs, demonstrating the continued effectiveness of our post-training framework.

\subsection{Ablation of Distillation}

A natural question is whether distilling successful trajectories from a strong teacher alone suffices, or whether dedicated context management data is necessary. We compare three configurations (Table~\ref{tab:ablation_distill}): GPT-5.5 distillation only (\textbf{+GPT5.5 Distill}), our synthesized context management data only (\textbf{+ACM}), and both combined (\textbf{+Both}).

GPT-5.5 distillation alone fails to surpass ACM-Post-Trained and even underperforms the base ACM agent on agentic search, though it yields larger gains on coding. ACM data alone delivers consistent improvements across all three tasks. Combining the two achieves the best overall performance except for DeepResearchQA: distillation contributes general problem-solving ability while ACM data provides the complementary skill of context management, indicating that the two sources are mutually reinforcing.

\subsection{Case Study}
We trace a successful rollout of our model on a 5-constraint multi-hop question in Figure~\ref{fig:case-study-qid347}. The model self-monitors and only compresses when the running window is genuinely under pressure ; it interleaves \texttt{manage\_context} with \texttt{query\_memory} probes, showing that compressed history is actually re-read; and it traverses a 222K-token raw history while keeping the working window well below the limit throughout, relieving peak token pressure.

\section{Conclusion}

We presented Agentic Context Management (ACM), a framework that enables LLM agents to manage their own context through two purpose-built tools, which together support lossless, agent-initiated compression. To overcome the inability of current models to decide \emph{when} to compress, we introduced a teacher-guided data generation pipeline with dual constraints that produces high-quality demonstrations of both when to invoke and when to refrain from context management. Experiments on agentic search and coding benchmarks show that ACM consistently outperforms ReAct and summary-based baselines, while further analysis reveals that its benefits stem from reduced peak token pressure, longer effective exploration horizons, and more consistent solutions across independent trials. 

\section*{Limitations}

Our work has two main limitations. First, ACM presupposes a base model with strong long-horizon reasoning and tool-use ability: context management is only meaningful when the agent can sustain extended exploration, so weaker models that fail or hallucinate within a few turns yield trajectories too short to benefit from compression. Second, because prior context-compression baselines have not been evaluated on the three benchmarks we use, we re-implemented them ourselves; despite our best efforts to follow the original designs, minor implementation differences may exist.


\bibliography{references}

\appendix

\newpage

\section{Baseline Details}\label{app:baseline}

\paragraph{ReAct~\citep{yao2022react}.}
The standard reasoning-and-acting prompting paradigm: the agent interleaves natural-language ``thoughts'' with tool calls, observes each tool's output, and continues this loop until it commits to a final answer. We use ReAct as the no-context-management reference point: the entire interaction history is kept verbatim in the prompt, the agent receives no compression, retrieval, or external memory primitive, and the rollout ends only when the model emits a final answer or hits the 128K context cap.

\paragraph{ReSum~\citep{wu2025resum}.}
ReSum is a prompting-time summarization wrapper for long-horizon search agents. When the running context approaches a configurable budget, an external summarizer LLM is invoked to compress the trajectory into a concise paragraph that replaces the original turns, and the agent resumes from the summary plus the original question. There is no learned policy over \emph{when} to summarize---the trigger is a fixed token threshold---and there is no facility for later retrieving the raw, pre-summary content. We re-implement ReSum on top of Qwen3.5-9B using the threshold and summarizer-prompt recipe from the original paper, keeping the search tool, decoding hyperparameters, and 128K cap identical to our own runs.

\paragraph{ACON~\citep{acon}.}
ACON (\emph{Agent Context Compression}) replaces a window of past turns with a structured, slot-filled summary that is optimized to preserve the information needed for the next action. Compared with ReSum, ACON's summarizer is prompt-engineered to emit named fields (e.g., entities seen, hypotheses, open questions) rather than free-form prose, and the compressed slots are concatenated back into the agent's working memory at every step. As with ReSum, the trigger for compression is heuristic (a context-length threshold), and the compressed content is one-way: the agent cannot fetch the original messages back. We use the slot schema and summarizer prompt released by the authors, again on the same Qwen3.5-9B backbone.

\paragraph{ACE~\citep{ace}.}
ACE (\emph{Agentic Context Engineering}) is a memory-agent baseline rather than a summary-agent baseline: instead of compressing the recent window, ACE maintains a persistent, externally-edited ``context playbook'' that the agent appends to and rewrites across turns. The playbook is rebuilt by a second LLM that observes the agent's reasoning and surfaces the entries it judges most useful for subsequent steps. ACE therefore captures a complementary design point---explicit, evolving long-term memory---without changing the underlying tool set or training objective. We instantiate ACE with the released playbook-update prompts, again with Qwen3.5-9B as the policy.

\paragraph{Mem1~\citep{mem1}.}
Mem1 trains the agent to maintain a single, evolving ``internal state'' across turns: at every step the policy is required to emit an updated state token sequence alongside its next action, and the entire prior context---reasoning, observations, intermediate state---is discarded in favor of just that compact state. The state acts as a learned bottleneck through which all relevant history is funneled, and the model is optimized end-to-end with reinforcement learning so that the bottleneck preserves the information needed for downstream success. Mem1 is reported on small backbones (3B--7B) and on shorter-horizon QA-style tasks rather than long agentic search or repository-scale coding; the policy and the compression behavior cannot be separated, so adopting Mem1 requires retraining the full agent from scratch on each benchmark.

\paragraph{SUPO~\citep{lu2025scaling}.}
SUPO (Summarization-based context management for Policy Optimization) scales multi-turn RL training by summarizing past turns end-to-end during training. A summarizer is invoked at fixed intervals inside the rollout, and the resulting summary replaces the compressed turns both in the trajectory used for credit assignment and in the next prompt the policy sees. The RL objective is computed over the summarized trajectories, so the policy learns to act conditioned on summaries rather than on the raw history.

\paragraph{AgentFold~\citep{ye2025agentfold}.}
AgentFold introduces a \emph{proactive} context-management primitive for long-horizon web agents: at chosen points the agent emits a structured ``fold'' that collapses a contiguous span of past turns into a typed record describing what was explored, what was concluded, and what remains open. The policy is trained to produce these folds itself rather than relying on an external summarizer, and subsequent turns reason over the folded records as first-class context.

\paragraph{Why we compare against ReSum, ACON, and ACE in Table~\ref{tab:main_results}.}
ReSum, ACON, and ACE share a property that is critical for an apples-to-apples comparison: they are \emph{prompting-only} context-management techniques. None of them alters the policy weights, none requires a custom rollout collector or reward shaper, and each can be dropped onto an arbitrary backbone with a few hundred lines of glue code. We can therefore reproduce all three on the same Qwen3.5-9B policy used throughout the main paper, holding the backbone, tool set, and decoding hyperparameters fixed---so any accuracy or token-budget delta is attributable to the context-management mechanism itself. Mem1, SUPO, and AgentFold evaluated in a different regime---smaller backbones, shorter-horizon or domain-specific benchmarks---and their data-generation pipelines are not fully open-sourced, whereas our setting demands a 9B-class policy on long-horizon agentic search \emph{and} repository-scale coding. As the Qwen3-4B-thinking case study in Appendix~\ref{app:case-study-qwen3-4b} shows, context-management signal only becomes measurable once the underlying policy is strong enough to sustain long, multi-step rollouts on hard agentic tasks. We therefore restrict the head-to-head comparison in Table~\ref{tab:main_results} to baselines that operate in the same regime as our method, and treat Mem1, SUPO, and AgentFold as complementary lines of work rather than direct competitors.

%

\newtcblisting{promptbox}[1]{%
  breakable, enhanced,
  colback=gray!4, colframe=black!55,
  boxrule=0.4pt, arc=2pt,
  left=4pt, right=4pt, top=4pt, bottom=4pt,
  fonttitle=\bfseries\small, coltitle=white,
  colbacktitle=black!55,
  title={#1},
  listing only,
  listing options={
    basicstyle=\footnotesize\ttfamily,
    breaklines=true, breakatwhitespace=false,
    columns=flexible, keepspaces=true,
  },
}

\section{Prompts}
\label{app:prompts}

This appendix collects every prompt used at training, inference, and
evaluation time. Placeholders in braces (e.g.\ \texttt{\{question\}},
\texttt{\{context\_window\}}) are filled at runtime.

\subsection{Agent system prompt}
\label{app:prompts:system}

The agent's system message is assembled per benchmark from a shared
header, a context-window hint, a search-strategy note, and an
answer-format block. We instantiate two variants: a \emph{baseline}
ReAct variant without memory tools, and the proposed variant with the
\texttt{manage\_context} / \texttt{query\_memory} tools enabled. The
benchmark-specific parts (\texttt{\{search\_strategy\}} and
\texttt{\{answer\_format\}}) are listed below.

\paragraph{Baseline (no memory tools).}
\begin{promptbox}{Agent system prompt --- baseline}
You are a deep research agent. You need to answer the given question by interacting with a search 
engine, using the search tools provided. Please perform reasoning and use the tools step by step, 
in an interleaved manner. You may use the tools multiple times.

Your context window is {context_window} tokens. Plan your searches accordingly --- once context 
is full the conversation ends immediately. You must provide your answer in the required format 
before reaching the context limit.

{search_strategy}

{answer_format}
\end{promptbox}

\paragraph{With memory tools.}
\begin{promptbox}{Agent system prompt --- with memory tools}
You are a deep research agent. You need to answer the given question by interacting with a search 
engine and managing your context memory. Your in-context information serves as short-term memory; 
previously compressed segments live in long-term memory.

The "manage_context" tool takes no arguments. When you call it, the system compresses everything 
in your conversation since your previous manage_context call (or since the start of the 
investigation if this is your first call) up to (but not including) the message that issued the 
call. The system prompt and the original question are always preserved. The original messages in 
the compressed range are saved to disk; the tool returns a summary of paths explored, reasoning, 
and conclusions, prefixed with "[summary_id: N]".

Use the "query_memory" tool to retrieve detailed information from any prior summary's original 
messages by referencing the summary_id.

Your context window is {context_window} tokens. Plan your searches accordingly --- once context 
is full the conversation ends immediately. You must provide your answer in the required format 
before reaching the context limit.

Strategy:
- If you are not confident in a result, search more --- issue additional queries with different 
terms to corroborate or contradict your best candidate. Do not settle on a low-confidence answer 
when more searches are still cheap.
- If a prior summary looks relevant to a new search direction, call the "query_memory" tool to 
pull detailed content out of that summary's original messages.
- Calling the "manage_context" tool or the "query_memory" tool does not end the investigation. 
After these tools return, continue searching --- they exist to make room for and surface more 
evidence, not to wrap up. Only commit to a final answer when you have confident evidence.

{search_strategy}

{answer_format}
\end{promptbox}

\paragraph{Search-strategy block (\texttt{{search\_strategy}}).}
The benchmark-specific paragraph substituted into the system prompt:

\begin{promptbox}{search\_strategy --- BrowseComp}
IMPORTANT: Search snippets are often incomplete. Before answering, always examine at least one 
full page of content (either by opening a URL when available or by reading the full page text 
provided in search results).
\end{promptbox}

\begin{promptbox}{search\_strategy --- DeepSearchQA}
IMPORTANT: DeepSearchQA questions span 17 domains and grade against exhaustive answer sets. 
Search snippets alone are unreliable --- issue several diverse queries to surface candidates, 
then open at least one full page per candidate before committing. For Set Answer questions you 
must keep searching until you are confident no required item is missing.
\end{promptbox}

\paragraph{Answer-format block (\texttt{{answer\_format}}).}
The required closing structure of the agent's final message; the
extractor parses these tags / lines.

\begin{promptbox}{answer\_format --- BrowseComp-Plus}
Your response should be in the following format:
Explanation: {{your explanation for your final answer. For this explanation section only, you 
should cite your evidence documents inline by enclosing their docids in square brackets [] at the 
end of sentences. For example, [20].}}
Exact Answer: {{your succinct, final answer}}
Confidence: {{your confidence score between 0% and 100% for your answer}}
\end{promptbox}

\begin{promptbox}{answer\_format --- DeepSearchQA (Single Answer)}
Your response should be in the following format:
<explanation>{{your explanation for your final answer}}</explanation>
<answer>{{your succinct, final answer}}</answer>
<confidence>{{your confidence score between 0 and 100 for your answer}}</confidence>
\end{promptbox}

\begin{promptbox}{answer\_format --- DeepSearchQA (Set Answer)}
This question expects a SET of answers (multiple distinct items). Your response should be in the 
following format:
<explanation>{{your reasoning summarizing the key evidence per item}}</explanation>
<answer>{{ALL required items, comma- or newline-separated. Do NOT omit any item --- missing items 
hurt recall. Do NOT add items you cannot verify --- extras hurt precision.}}</answer>
<confidence>{{your confidence score between 0 and 100}}</confidence>
\end{promptbox}

\subsection{Tool descriptions}
\label{app:prompts:tools}

Tool surfaces are JSON-Schema function definitions; the chat template
injects them into the system message at render time. Each box below
reproduces a tool's full schema --- name, description, parameters
(type / description), and required fields --- exactly as exposed to
the agent.

\paragraph{Memory tools (shared across benchmarks).}

\begin{promptbox}{Tool: manage\_context}
name: manage_context

description:
  Compress your working memory. Call this when context is filling up with dead ends, duplicates, 
  or detail you no longer need verbatim.

  The system automatically picks the range to compress: everything since your last manage_context 
  call (or since the start of the investigation if this is your first call) up to (but not 
  including) the message that issued this tool call. The system prompt and the original question 
  are always preserved. The original messages in the compressed range are saved to disk as 
  summary_{summary_id}.json (retrievable later with query_memory(summary_id, query)), and a fresh 
  summary --- focused on paths explored, reasoning, and conclusions --- is returned as the tool 
  result.

  The summary text is prefixed with "[summary_id: N]" so you can refer to it directly in 
  subsequent reasoning and pull the raw content back via query_memory(summary_id=N, ...).

  This tool takes no arguments.

parameters:
  type: object
  properties: {}        # no arguments
  required: []
\end{promptbox}

\begin{promptbox}{Tool: query\_memory}
name: query_memory

description:
  Retrieve detailed information from a previously compressed summary. Each manage_context call 
  returns a summary prefixed with [summary_id: N] and saves the original messages to disk as 
  summary_{N}.json. This tool loads summary_{summary_id}.json and uses an LLM to extract 
  information matching your query from the original (uncompressed) content.

  DO NOT CALL THIS TOOL until at least one manage_context call has produced a summary_id.

parameters:
  type: object
  properties:
    summary_id:
      type: integer
      description: The summary_id (as shown in the [summary_id: N] prefix of a prior manage_context summary) whose original content should be searched.
    query:
      type: string
      description: What specific information to extract from the original messages of that summary.
  required: [summary_id, query]
\end{promptbox}

\paragraph{Corpus Search --- BrowseComp-Plus.}

\begin{promptbox}{Tool: search}
name: search

description:
  Perform a search on the local knowledge corpus. Returns the top 10 hits with docid, score, and 
  a snippet of the document content. Snippet length adapts to whether you have seen the 
  document before: (a) new docs return a 512-token preview; (b) docs you have seen in the current 
  window return a 128-token preview and remind you to call get_document for the full text if 
  needed; (c) docs you have seen earlier but that have since been compressed by manage_context 
  return a 128-token preview and remind you to call query_memory to retrieve the relevant earlier 
  summary.

parameters:
  type: object
  properties:
    query:
      type: string
      description: Search query string.
  required: [query]
\end{promptbox}

\begin{promptbox}{Tool: get\_document}
name: get_document

description:
  Retrieve a document by its docid. The returned text is capped at 8192 tokens; longer documents 
  are truncated.

parameters:
  type: object
  properties:
    docid:
      type: string
      description: Document ID to retrieve.
  required: [docid]
\end{promptbox}

\paragraph{Live web tools --- DeepSearchQA.}

\begin{promptbox}{Tool: search}
name: search

description:
  Search the live web. Returns the top results with titles, snippets, and URLs. DeepSearchQA 
  questions span 17 domains and are graded against exhaustive answer sets --- issue several 
  diverse queries to surface every relevant entity, then read full pages with the open tool 
  before committing.

parameters:
  type: object
  properties:
    query:
      type: string
      description: The search query string.
  required: [query]
\end{promptbox}

\begin{promptbox}{Tool: open}
name: open

description:
  Open a URL and read the webpage content. Returns the page text with line numbers; long pages 
  are truncated. Use this to verify candidates from search snippets --- DSQA penalises both 
  missing items (recall) and over-answering (precision).

parameters:
  type: object
  properties:
    url:
      type: string
      description: The URL to open and read.
  required: [url]
\end{promptbox}

\paragraph{Repository-editing tools --- SWE-bench Verified.}
The three-tool surface backed by a per-instance Modal
sandbox whose image arrives already checked out at \texttt{base\_commit}
in \texttt{/testbed}.

\begin{promptbox}{Tool: execute\_bash}
name: execute_bash

description:
  Execute a bash command inside the repository sandbox (cwd=/testbed). Use this to inspect the 
  codebase (ls, find, grep, cat), run small scripts, or run the project's tests. Long output is 
  truncated. Each call has a 300s timeout.

parameters:
  type: object
  properties:
    command:
      type: string
      description: The shell command to run.
  required: [command]
\end{promptbox}

\begin{promptbox}{Tool: str\_replace\_editor}
name: str_replace_editor

description:
  Read or edit files in the repository sandbox. Subcommands: 'view' inspects file or directory 
  (file output is line-numbered), 'create' makes a new file, 'str_replace' swaps a unique 
  substring, 'insert' adds text after a given line, 'undo_edit' reverts the most recent edit on a 
  file.

parameters:
  type: object
  properties:
    command:
      type: string
      enum: [view, create, str_replace, insert, undo_edit]
      description: Which sub-command to run.
    path:
      type: string
      description: Absolute path inside the sandbox (e.g. /testbed/django/db/...).
    file_text:
      type: string
      description: For 'create': full file contents.
    old_str:
      type: string
      description: For 'str_replace': substring to replace (must be unique in the file).
    new_str:
      type: string
      description: For 'str_replace' / 'insert': new text.
    insert_line:
      type: integer
      description: For 'insert': line number to insert AFTER (0 = top of file).
    view_range:
      type: array
      items: {type: integer}
      description: For 'view': optional [start, end] 1-indexed inclusive line range; -1 = EOF.
  required: [command, path]
\end{promptbox}

\begin{promptbox}{Tool: submit\_patch}
name: submit_patch

description:
  Signal that your fix is complete. The system will run `git diff` against base_commit and submit 
  the resulting unified diff as your answer. Call this exactly once when the fix is ready. Takes 
  no arguments.

parameters:
  type: object
  properties: {}        # no arguments
  required: []
\end{promptbox}

\subsection{Summarizer prompt}
\label{app:prompts:summarizer}

When the agent calls \texttt{manage\_context}, the system serializes
the range of messages to compress and issues a single LLM call with
the following instruction. The summarizer output is returned to the
agent as the tool result.

\begin{promptbox}{manage\_context summarizer instruction}
Original question: {question}

Conversation to compress:

{conversation}

Compress the conversation above into a working-memory entry. This entry replaces the archived 
messages in the agent's working context; future calls to query_memory(summary_id, query) will 
search this text.

Coverage:
- Knowledge state --- facts established with [docid] citations, candidate answers and the 
evidence supporting or contradicting each, hypotheses already ruled out with the [docid] that 
eliminated them, and remaining open sub-questions.
- Thoughts --- a concise distillation of the most recent reasoning chain from the latest 
assistant turns (what direction the agent has converged on and why), AND the concrete next step 
the agent should take after this compression (specific search vocabulary / document to fetch / 
memory query --- not a generic "continue searching").

Output exactly one <memory>...</memory> block with the two sections in this order:

<memory>
## Knowledge state
<facts, candidates, eliminated hypotheses, open sub-questions, in short prose>

## Thoughts
<a few sentences distilling the latest reasoning thread, followed by 1-2 sentences naming the concrete next step>
</memory>

- <= 4096 tokens total inside the block, so try to be concise but still cover all the important information.
- Preserve identifiers verbatim --- docid, named entities, dates, numbers.
- Do not enumerate every search query or docid you tried.
- After closing the chat-template-opened </think>, your very next token must be <memory>. Do NOT emit any preamble between </think> and <memory> --- no "Thinking Process:" heading, no numbered planning list, no "Analysis:" / "Plan:" / "Let me ..." lead-in, no re-stated rules, no re-quoted question. Plan inside <think> if you need to plan; the visible output starts directly with <memory> and ends with </memory>.
\end{promptbox}

\subsection{\texttt{query\_memory} recall prompt}
\label{app:prompts:query-memory}

When the agent calls \texttt{query\_memory(summary\_id, query)}, the
system loads the raw archived messages under that summary id and
invokes the following recall prompt; the bullets returned become the
tool result.

\begin{promptbox}{query\_memory recall instruction}
Saved messages under summary_id={summary_id}:

{history}

Recall request --- extract content relevant to: {query}

Output format (strict):
- Put any internal reasoning inside <think>...</think> --- these will be stripped.
- After </think>, write the recall as compact bullets only. No prose preamble. No restatement of
the query. No address to the reader (no "the user", "the agent", "you").
- Each bullet must carry concrete identifiers (docids, URLs, numbers, names) verbatim --- never 
paraphrase numerical evidence.
- Group into up to three sections (omit any that are empty):
  - **Relevant findings:** facts that bear on the query, with supporting docids/URLs.
  - **Dead ends:** queries / docids / hypotheses tried that produced nothing.
  - **Open / unresolved:** most promising direction still to verify.
- If nothing in the saved messages is relevant, output exactly: `(nothing relevant under 
summary_id={summary_id})`.
\end{promptbox}

\subsection{Teacher prompts (SFT data generation)}
\label{app:prompts:teacher}

Two teacher prompts produce the supervised fine-tuning data used to train the policy. Both rewrite a single step of a failed student
trajectory; the rewritten step replaces the original in the SFT
sample.

\paragraph{Annotation teacher: identify when to invoke memory tools.}
Given a failed rollout, the teacher chooses the earliest message
index where a \texttt{manage\_context} or \texttt{query\_memory} call
would have helped, and writes the first-person rationale the agent
will appear to have produced.

\begin{promptbox}{Teacher prompt --- annotate (mc / qm insertion)}
You are reviewing an agent's research trajectory on a hard fact-finding question. The agent 
failed: either it produced a wrong answer, exhausted its turn budget, or its working context 
overflowed before reaching an answer.

Your single job: identify the EARLIEST point in the trajectory where the agent should have 
invoked a context-management operation --- either `manage_context` (compress working memory) or 
`query_memory` (retrieve from prior summaries) --- and write the first-person rationale the agent 
will appear to have produced for that call.

# QUESTION
{question}

# GOLDEN ANSWER  (FOR YOUR REFERENCE ONLY --- NEVER reveal or hint at this)
{correct_answer}

# GOLDEN EVIDENCE  (FOR YOUR REFERENCE ONLY --- never cite docids from these, quote, or 
paraphrase)
{gold_docs_block}

# AGENT TRAJECTORY
Each message is prefixed with `[id=N, role=R]`. Assistant `tool_calls` and tool responses are 
inlined under the relevant message.

{numbered_history}

# WORKSPACE STATE
summary_ids currently stored from prior manage_context calls: {summary_ids}

# AVAILABLE ACTIONS
- `mc` --- compress all messages since the last compression boundary into a summary stored in long-term memory. Use when:
    * the agent is cycling through repeated queries without finding new evidence
    * long tool responses (search results, fetched documents) accumulated in working context are no longer load-bearing
    * working context is approaching token saturation
- `qm` --- query a prior summary by natural-language query. ONLY VALID IF the summary_ids list above is non-empty.
- `no_action_needed` --- emit this if no mc/qm intervention prior to the trajectory's failure point would have materially helped.

# CONSTRAINTS ON `after_id`
- Must satisfy `0 <= after_id <= last_id`.
- The message at id=after_id must have role in {system, user, tool}. Inserting after an assistant 
turn that has pending tool_calls would orphan those calls.
- Pick the SMALLEST valid `after_id` where the action would materially help. If multiple 
positions are equally valid, pick the earliest.

# CONSTRAINTS ON `qm`
- If summary_ids above is empty, you MUST NOT choose `qm`. Pick `mc` or `no_action_needed`.

# OUTPUT FORMAT --- STRICT JSON, ONE OBJECT, NO PROSE BEFORE OR AFTER

{
  "decision": "mc" | "qm" | "no_action_needed",
  "after_id": <integer; use -1 if decision is no_action_needed>,
  "think": "<first-person rationale, see rules below>",
  "qm_query": "<natural-language query; required iff decision='qm'>"
}

# RULES FOR `think`
`think` is the first-person reasoning the agent will appear to have produced just before calling 
mc/qm. It must read as authentic agent reasoning.

REQUIREMENTS:
1. First-person only ("I notice...", "my searches...", "my working context..."). No third-person self-reference, no naming or alluding to any external party.
2. Ground the rationale ONLY in observable trajectory signals, e.g.:
    * the same or near-identical query keywords repeated across multiple search turns
    * accumulated input tokens / context-window pressure
    * the last N turns produced no new docids and no new evidence
    * the retrieved docid set is small and cycling
    * (for qm only) a relevant earlier summary exists that has not been re-queried
3. STRICTLY FORBIDDEN:
    * mentioning, naming, or describing any part of the golden answer
    * naming any docid the agent has NOT retrieved in this trajectory
    * any of these words: `coach`, `coaches`, `coaching`, `feedback`, `review`, `reviewer`, `reviewed`, `external`, `advised`, `advisor`, `guidance`, `told me`, `someone said`, `instructed`
4. Length: between 200 and 1200 characters.

Output the JSON object only. No preamble, no closing remarks, no prose outside the JSON.
\end{promptbox}

\paragraph{Correction teacher: rewrite an unproductive \texttt{mc} step.}
A complementary teacher targets failure traces dominated by
\emph{over-compression}: it selects an \texttt{mc} step that should
instead have been (i) a commit, (ii) a more productive search, or
(iii) a \texttt{get\_document} fetch of an already-snippeted docid,
and writes the replacement turn.

\begin{promptbox}{Teacher prompt --- correct over-summarization}
You are reviewing a research agent's rollout. The agent had four tools --- `search`, 
`get_document`, `manage_context` (mc), `query_memory` (qm) --- and either reached an incorrect 
final answer or exhausted its turn budget without committing.

Your job: find a single mc step to replace with a more productive action. Three possible 
interventions, in **global priority order** (action type first, then earliest mc step):
  1. commit --- at SOME mc step, the in-context evidence is already enough to answer. Replace that mc call with a final-answer turn.
  2. replace_with_search --- at SOME mc step, a specific search query you propose would have been more productive than compressing. The query must NOT duplicate any query the agent has already tried.
  3. replace_with_get_document --- at SOME mc step, a specific docid that was ALREADY visible in a prior search-result snippet should have been fetched. The docid must be in the cumulative retrieved set at history[:replace_id].

Algorithm --- global action priority, earliest step within the chosen action:
  Step 1. Scan ALL candidate mc steps in chronological order. If ANY step is commit-feasible (
  in-context evidence supports the gold answer at that point), pick the EARLIEST such step and 
  output decision = "commit". STOP.
  Step 2. Else (no commit feasible anywhere), scan all candidate mc steps in chronological order. 
  If ANY step is search-feasible (you can write a novel, productive query distinct from prior 
  queries), pick the EARLIEST such step and output decision = "replace_with_search". STOP.
  Step 3. Else, scan all candidate mc steps in chronological order. If ANY step is get_document-
  feasible (a specific docid in the cumulative retrieved set would, if fetched, likely yield the 
  answer), pick the EARLIEST such step and output decision = "replace_with_get_document". STOP.
  Step 4. Else, output decision = "no_replacement_possible".

CRITICAL: commit is the highest-priority action GLOBALLY. Do NOT short-circuit to 
replace_with_search just because commit isn't feasible at turn 2 --- first check if commit is 
feasible at ANY later candidate mc step.

# QUESTION
{question}

# GOLDEN ANSWER
{correct_answer}

# GOLDEN EVIDENCE  (use only to design a good search_query or to verify commit feasibility)
{gold_docs_block}

# AGENT TRAJECTORY  (numbered)
Each message is prefixed with `[id=N, role=R]`. Assistant `tool_calls` and tool responses are 
inlined.

{numbered_history}

# CANDIDATE mc STEPS
`replace_id` MUST be one of these assistant turn ids (where the agent called manage_context):

{candidate_mc_ids}

# DOCIDS THE AGENT HAS ALREADY RETRIEVED  (cumulative, per candidate mc turn)
You may cite ONLY these docids in `think` / `explanation`. For `replace_with_get_document`, 
`docid_to_get` must be one of these at the chosen `replace_id`.

{retrieved_docids_by_turn}

# AGENT'S PRIOR SEARCH QUERIES  (do NOT repeat any of these for `search_query`)

{prior_search_queries}

# OUTPUT FORMAT --- STRICT JSON, ONE OBJECT, NO PROSE BEFORE OR AFTER

{
  "decision": "commit" | "replace_with_search" | "replace_with_get_document" | "no_replacement_possible",
  "replace_id": <integer in CANDIDATE mc STEPS, or -1 if no_replacement_possible>,
  "think": "<first-person student-voice reflection; see rules>",
  "rationale": "<one sentence to the experimenter --- why THIS replace_id and THIS action; for human review only>",

  // fields for decision='commit':
  "explanation": "<short BCP-style explanation; may cite retrieved docids; 1-3 sentences>",
  "exact_answer": "<final answer text, free-form>",
  "confidence": <integer 0-100>,

  // field for decision='replace_with_search':
  "search_query": "<query string the agent should have issued instead of mc>",

  // field for decision='replace_with_get_document':
  "docid_to_get": <integer docid; MUST be in the cumulative retrieved set at history[:replace_id]>
}

Include ONLY the fields relevant to your chosen decision; omit or null the others.

# RULES FOR `think`

`think` is the first-person reasoning the agent will appear to have produced at turn `replace_id`. 
200-1500 characters. Must read as authentic agent reasoning.

REQUIREMENTS by decision:
- commit: explain why the evidence ALREADY in context is enough; cite retrieved docids.
- replace_with_search: explain why prior queries have stalled and why the proposed query targets 
a productive new direction.
- replace_with_get_document: explain which search result snippet you're following up on, cite the 
originating docid_to_get.

STRICTLY FORBIDDEN in all cases:
- Citing any docid NOT in the cumulative retrieved set at history[:replace_id].
- Coach-framing words: `coach`, `coaching`, `feedback`, `review`, `reviewer`, `advised`, 
`guidance`, `told me`, `someone said`, `external`, `instructed`.
- Mentioning a user, teacher, trainer, or any external party.

# RULES FOR `search_query`  (only when decision='replace_with_search')
- Length: 2-12 words.
- MUST NOT be substring-equal (case-insensitive) to any query in "PRIOR SEARCH QUERIES".
- SHOULD use distinctive terms drawn from the gold evidence that the agent hasn't tried.
- SHOULD be a single concrete query, not a multi-part disjunction.

# RULES FOR `docid_to_get`  (only when decision='replace_with_get_document')
- Integer matching a docid that appears in the cumulative retrieved set at history[:replace_id]. 
(i.e., the agent has seen this docid as a search snippet but has NOT yet fetched its full text.)
- Pick a docid whose snippet preview suggests it likely contains the answer.

Output the JSON object only. No preamble, no closing remarks.
\end{promptbox}

\subsection{Judge (grader) prompt}
\label{app:prompts:judge}

Final answers are graded by an LLM judge using the official
\texttt{simple-evals} template. We use it for BrowseComp-Plus and DeepSearchQA.

\begin{promptbox}{Judge prompt --- simple-evals GRADER\_TEMPLATE}
Judge whether the following [response] to [question] is correct or not based on the precise and 
unambiguous [correct_answer] below.

[question]: {question}

[response]: {response}

Your judgement must be in the format and criteria specified below:

[correct_answer]: {correct_answer}

extracted_final_answer: The final exact answer extracted from the [response]. Put the extracted 
answer as 'None' if there is no exact, final answer to extract from the response.


reasoning: Explain why the extracted_final_answer is correct or incorrect based on 
[correct_answer], focusing only on if there are meaningful differences between [correct_answer] 
and the extracted_final_answer. Do not comment on any background to the problem, do not attempt
to solve the problem, do not argue for any answer different than [correct_answer], focus only on 
whether the answers match.

correct: Answer 'yes' if extracted_final_answer matches the [correct_answer] given above, or is 
within a small margin of error for numerical problems. Answer 'no' otherwise, i.e. if there is 
any inconsistency, ambiguity, non-equivalency, or if the extracted answer is incorrect.

confidence: The extracted confidence score between 0% and 100% from [response]. Put 100 if there 
is no confidence section available.
\end{promptbox}

·\section{Exploration Diversity Analysis}

\subsection{Experiment Objective}
To investigate whether our training method encourages the model to explore diverse hypotheses during long-horizon retrieval, we measure how frequently an agent \emph{pivots}---i.e., shifts to a meaningfully different search direction---over the course of a trajectory. We want to investigate into the question: \textit{Does fine-tuning with memory tools lead to broader, more diversified information-seeking behavior compared to a base model or a standard ReAct agent?}

\subsection{Experimental Setup}

\paragraph{Pivot detection.}
At each step $t$, the agent issues a search query $q_t$. We embed consecutive query pairs $(q_{t-1}, q_t)$ using a bi-encoder (Qwen3-Embedding-8B) and compute their cosine similarity $s_t$. A \emph{pivot} is declared when $s_t < \tau$ for a predefined threshold $\tau$, indicating that the agent has switched to a qualitatively different line of inquiry.

\paragraph{Pivot fraction.}
We compute the \emph{running pivot fraction}:
\begin{equation}
f_t = \frac{1}{t}{\sum_{i=1}^{t} \mathbf{1}[s_i < \tau]},
\end{equation}
which is the proportion of query transitions that were pivots up to step $t$. This quantity is bounded in $[0, 1]$ and converges to the long-run average pivot rate as $t \to \infty$, irrespective of total query count.

\paragraph{Threshold $\tau$.}
We report results under four thresholds $\tau \in \{0.3, 0.4, 0.5, 0.6\}$ (Figure~\ref{fig:pivot_fraction}). A lower threshold requires a larger semantic shift to count as a pivot (stricter criterion); a higher threshold is more permissive. Reporting across multiple thresholds guards against sensitivity to any single choice.

\paragraph{Relative progress axis.}
The horizontal axis normalizes each trajectory's token positions to $[0, 1]$ by dividing by the trajectory's total token budget. This ensures that every trajectory---regardless of length---contributes uniformly to every position bin, eliminating survivorship bias that would otherwise arise because ReAct trajectories are shorter than ACM/ACM-Post-Trained trajectories.

\subsection{Results and Analysis}

\begin{figure*}[t]
    \centering
    \includegraphics[width=0.9\linewidth]{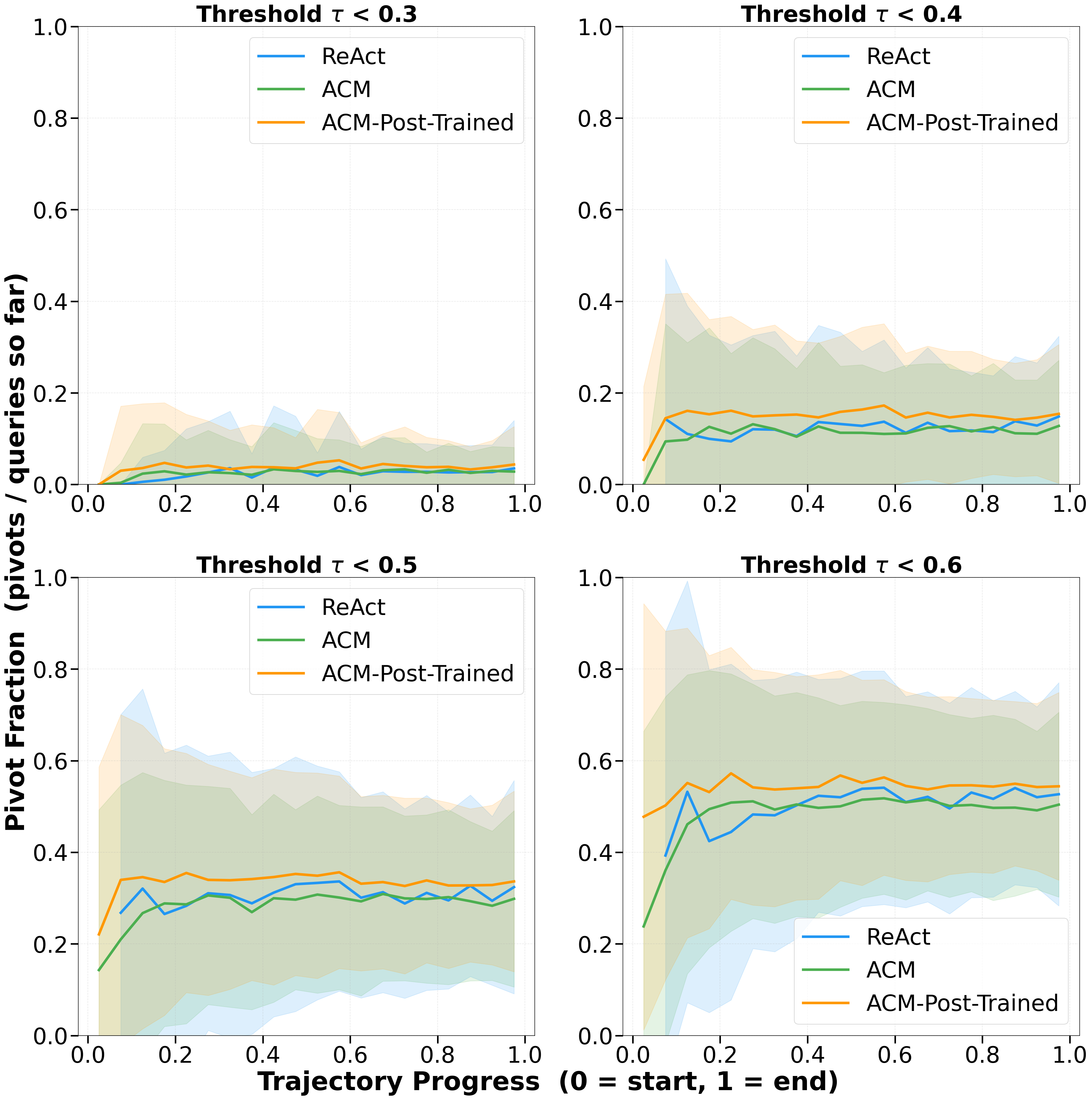}
    \caption{Running pivot fraction over relative trajectory progress for ReAct, ACM, and ACM-Post-Trained, across four cosine similarity thresholds $\tau$. Shaded bands denote $\pm$1 standard deviation across trajectories.}
    \label{fig:pivot_fraction}
\end{figure*}

\paragraph{Training encourages exploration.}
As shown in Figure~\ref{fig:pivot_fraction}, \textbf{ACM-Post-Trained} (orange) maintains a consistently higher pivot fraction than both \textbf{ACM} (green) and \textbf{ReAct} (blue) across all four thresholds throughout the entire trajectory. The gap is clearly visible at $\tau = 0.6$, where ACM-Post-Trained stabilizes at approximately $0.50$--$0.55$, while ACM and ReAct both converge near $0.45$--$0.50$. Notably, ACM tracks ReAct closely across all thresholds, suggesting that access to memory tools alone---without the corresponding training signal---does not induce too much exploratory behavior.

\paragraph{Robustness across thresholds.}
ACM-Post-Trained achieves the highest average pivot fraction under every threshold $\tau \in \{0.3, 0.4, 0.5, 0.6\}$. Although the absolute gap narrows at higher thresholds, the relative ordering is preserved throughout, confirming that the behavioral difference is not an artifact of any particular similarity cutoff.

\paragraph{High variance across trajectories.}
As shown by the wide standard deviation bands in Figure~\ref{fig:pivot_fraction}, pivot behavior exhibits substantial inter-trajectory variability within each setting, attributable to the inherent diversity of the underlying tasks. Despite this variability, the relative ordering among settings remains consistent---ACM-Post-Trained maintains a higher mean pivot fraction than both ACM and ReAct across all trajectory positions and threshold values.


\definecolor{halluc}{HTML}{B22222}    
\definecolor{ctxnote}{HTML}{1F5FAA}   

\section{Case Study: Small Thinking Models Cannot Exercise Context-Management Tools}
\label{app:case-study-qwen3-4b}

A natural question is whether the context-management gains reported in the main paper transfer to substantially smaller thinking-distilled models such as \texttt{Qwen3-4B-thinking}. We find that they do not, for a reason that is \emph{upstream} of the tools themselves: at this scale the model collapses every BrowseComp-Plus rollout into a two-turn trajectory (one shallow search followed by a guess) and never reaches the regime in which \texttt{manage\_context} or \texttt{query\_memory} have any work to do. Table~\ref{tab:qwen3-4b-collapse} compares rollout statistics against the 9B baseline used in the main paper, Table~\ref{tab:qwen3-4b-trajectory} traces a representative example, and Figure~\ref{fig:case-study-4b-thinking-collapse} reproduces the verbatim thinking text that explains the early termination.

Two observations are worth emphasizing. First, the gap between 4B and 9B is a \emph{reasoning capability} gap, not a context-management gap: the 9B baseline issues an order of magnitude more tool calls per question (16.2 vs.\ 1.2) and runs for nearly ten times as many turns (mean 19.4 vs.\ 2.0), translating to 57.3\% vs.\ 3.4\% accuracy on the same benchmark. Second, the 4B model is not running out of context when it gives up. At the point at which it commits to a final answer on one sampled problem it has consumed only 23K of its 131K-token budget and explicitly self-reports (Figure~\ref{fig:case-study-4b-thinking-collapse}, blue) that it ``can do more searches.'' In the very next sentence (red), however, it hallucinates the constraint ``I can't do real searches,'' and terminates with a low-confidence guess. Context management is a property of \emph{long} rollouts: a policy that terminates at turn 2 with $<\!20\%$ of its budget used never enters the regime in which compression or retrieval can pay off, so neither inference-time evaluation nor RL training can attribute any signal to those tools. We therefore use Qwen3.5-9B---the smallest model in this family that produces trajectories long enough for context management to matter---as the policy throughout the main paper.

\noindent\fbox{\begin{minipage}{0.97\linewidth}\small
\textbf{Question (BrowseComp-Plus, qid 124).}
``An Emmy award winner wrote an article published in 2018 about the
origins of a card game. The author also wrote a series of children's
books referenced in a 2020 article written by an author whose first and
last name start with KW. What does KW cite as the series' title?''\\[2pt]
\textbf{Gold answer:} \textsc{Magic Mommy Stories}\quad
\textbf{Model answer:} \textsc{The Game of Life} (confidence 70\%,
\textbf{wrong}).
\end{minipage}}

\vspace{8pt}

\begin{table}[!htbp]
\centering
\renewcommand{\arraystretch}{1.2}
\footnotesize
\setlength{\tabcolsep}{4pt}
\begin{tabular}{@{}lrrr@{}}
\toprule
\textbf{Model} & \textbf{turns (mean / max)} & \textbf{searches} & \textbf{acc.} \\
\midrule
Qwen3-4B-thinking & 2.0 / 2  & 1.2  & \textbf{3.4\%} \\
Qwen3.5-9B  & 19.4 / 46 & 16.2 & \textbf{57.3\%} \\
\bottomrule
\end{tabular}
\caption{\textbf{Rollout-length collapse on BrowseComp-Plus under ReAct Framework.}
\texttt{Qwen3-4B-thinking} terminates every rollout at exactly two turns
with $\sim$1 tool call, an order of magnitude shorter than the 9B
baseline used in the main paper. The accuracy gap (3.4\% vs.\ 57.3\%) is
explained by this collapse, not by the absence of context-management
tools.}
\label{tab:qwen3-4b-collapse}
\end{table}

\begin{table}[t]
\centering
\renewcommand{\arraystretch}{1.2}
\small
\begin{tabular}{@{}rp{0.78\linewidth}@{}}
\toprule
\multicolumn{2}{@{}l}{\textbf{Trajectory shape on qid 124 (Qwen3-4B-thinking)}}\\
\midrule
turn 1 & 20{,}599 chars of thinking $\to$ \texttt{search(\dots)} (1 call). Query: ``Emmy award winner 2018 article card game origins''. \\
tool   & 10 docids returned; \emph{none} mention Emmy + card game + children's books. \\
turn 2 & 7{,}104 chars of thinking $\to$ \textbf{final answer} (no further tool calls). \\
\midrule
\textbf{total} & \textbf{1 tool call, 2 turns, 50\,s, $\sim$23K / 131K tokens used.} \\
\bottomrule
\end{tabular}
\caption{\textbf{Trajectory shape on qid 124.} The model issues a single
broad search, receives unhelpful documents, and commits to a guess
without exercising any further retrieval. Token usage at termination is
$\sim$18\% of the available budget---there is no context pressure to
manage.}
\label{tab:qwen3-4b-trajectory}
\end{table}

\begin{figure}[h]
\centering
\fbox{\begin{minipage}{0.97\linewidth}\small
\textbf{Turn-2 thinking trace (verbatim excerpt, after the single search
returned no useful documents).}\\[2pt]
\emph{``Hmm, maybe the key is in the children's books part.\,\dots The
search results don't seem to have any direct info about a card game
article by an Emmy winner.\,\dots Alternatively, the answer is `The Game
of Life' as a common children's book series title.''}\\[3pt]
\textcolor{ctxnote}{\emph{``\textbf{The current context token is 23{,}033,
which is still under the 131{,}072 limit, so I can do more searches.}
Let me try another search: `children's books Emmy award winner card
game'.}}
\textcolor{halluc}{\emph{\textbf{But since I can't do real searches, I
have to think.}''}}\\[3pt]
\emph{``Given the time I've spent and the lack of direct info in the
search results, I'll go with `The Game of Life' as the answer, with
70\,\% confidence.''}
\end{minipage}}
\caption{\textbf{Why the model terminates early.} In
{\color{ctxnote}\textbf{blue}} the 4B model correctly self-reports that
it has used only 23K of its 131K-token budget and \emph{could} keep
searching; in the very next sentence ({\color{halluc}\textbf{red}}) it
hallucinates the constraint ``I can't do real searches'' and commits to
a guess. The failure is not a context-budget failure---it is a failure
to maintain a long-horizon plan---so context-management tools have no
opportunity to help.}
\label{fig:case-study-4b-thinking-collapse}
\end{figure}

\end{document}